\documentclass[10p]{article} 
\usepackage{simpleConference}
\usepackage{times}
\usepackage{graphicx}
\usepackage{amssymb}
\usepackage{url,hyperref}
\usepackage[utf8]{inputenc} 
\usepackage[T1]{fontenc}    
\usepackage{hyperref}       
\usepackage{url}            
\usepackage{booktabs}       
\usepackage{amsfonts}       
\usepackage{nicefrac}       
\usepackage{microtype}      
\usepackage{amsmath}
\usepackage{tikz}
\usetikzlibrary{arrows}
\usepackage{amssymb}
\usepackage[normalem]{ulem}
\usepackage{soul,xcolor}
\usepackage{multirow,tabularx,booktabs,ctable}
\usepackage{subcaption}
\usepackage{xspace}
\usepackage{enumitem}
\usepackage{authblk}

\newcommand{\TAPAS}{TAPAS\xspace} 
\newcommand{\AP}{TAP\xspace} 
\newcommand{\DCC}{DC\xspace} 
\newcommand{\LCDB}{LDE\xspace} 
\newcommand{\DCN}{DCN\xspace}
\begin{document}

\title{TAPAS: Train-less Accuracy Predictor for Architecture Search}

\author[1,2]{R.\ Istrate}
\author[1]{F.\ Scheidegger}
\author[1]{G.\ Mariani}
\author[2]{D.\ Nikolopoulos}
\author[1]{C.\ Bekas}
\author[1]{A.\ C.\ I.\ Malossi}
\affil[1]{IBM Research -- Zurich, Switzerland}
\affil[2]{Queen's University of Belfast, United Kingdom}

\maketitle
\thispagestyle{empty}

\begin{abstract}
    In recent years an increasing number of researchers and practitioners have been suggesting algorithms for large-scale neural network architecture search: genetic algorithms, reinforcement learning, learning curve extrapolation, and accuracy predictors. None of them, however, demonstrated  high-performance without training new experiments in the presence of unseen datasets. We propose a new deep neural network accuracy predictor, that estimates in fractions of a second classification performance for unseen input datasets, without training. In contrast to previously proposed approaches, our prediction is not only calibrated on the topological network information, but also on the characterization of the dataset-difficulty which  allows us to re-tune the prediction without any training. Our predictor achieves a performance which exceeds 100~networks per second on a single GPU, thus creating the opportunity to perform large-scale architecture search within a few minutes. We present results of two searches performed in 400 seconds on a single GPU. Our best discovered networks reach 93.67\% accuracy for CIFAR-10 and 81.01\% for CIFAR-100, verified by training. These networks are performance competitive with other automatically discovered state-of-the-art networks however we only needed a small fraction of the time to solution and computational resources.
\end{abstract}

\section{Introduction}

Automatic generation and tuning of convolutional neural network (CNN) architectures is a growing research topic. The majority of approaches in the literature (for a deep overview, see Section~\ref{sec:related_works}) are rooted into the fundamental idea of large-scale explorations; more precisely, they can be based either on evolution and mutations~\cite{real2017large,DBLP:journals/corr/XieY17,DBLP:journals/corr/MiikkulainenLMR17}, or on reinforcement learning~\cite{DBLP:journals/corr/abs-1708-05552,DBLP:journals/corr/ZophL16,zoph2017learning,cai2018efficient,baker}. All these algorithms require a large amount of training experiments which quickly leads to massive resource and time to solution requirements.

Recently, the concept of performance prediction for architecture search has emerged. The fundamental idea is to drastically reduce exploration cost, by forecasting accuracy of networks without (or with very limited) training. Prediction is obtained either from partial learning curves~\cite{lce,bnn,svr,freeze}, or from a database of trained experiments~\cite{peephole}. The former approach requires partial training of each specific network. The latter one, implies training hundreds of networks on the given input dataset, to build a reliable ground-truth. Thus, none of them can be used out-of-the-box for near real-time architecture search. 

In this work, we introduce a train-less accuracy predictor for architecture search~(\TAPAS), that provides reliable architecture peak accuracy predictions when used with unseen (i.e., not previously seen by the predictor) datasets. This is achieved by adapting the prediction to the \emph{difficulty} of the dataset, that is automatically determined by the framework. In addition, we reuse experience accumulated from previous experiments. The main features of our framework are summarized as follows: (i)~it is not bounded to any specific dataset, (ii)~it learns from previous experiments, whatever dataset they involve, improving prediction over usage and (iii)~it allows to run large-scale architecture search on a single GPU device within a few minutes.

In summary, our main contributions are the following:

\begin{itemize}
\item a fast, scalable, and reliable framework for CNN architecture performance prediction;
\item a flexible prediction algorithm, that dynamically adapts to the \emph{difficulty} of the input; 
\item an extensive comparison with preexisting methods/results, clearly illustrating the advantages of our approach.

\end{itemize}

The outline of the paper is the following: in Section~\ref{sec:related_works} we briefly review literature approaches and analyze pros and cons of each of them. In Section~\ref{sec:methodology} we present the design of our prediction framework, with a deep dive into its three main components. Then, in Section~\ref{sec:experiments} we compare experimental results with current state-of-the-art. Finally, conclusions are summarized in Section~\ref{sec:conclusion}. 

\section{Related work}\label{sec:related_works}

This paper follows a similar design idea as Peephole~\cite{peephole}, which predicts a network accuracy by only analyzing the network structure.
Similar to our approach, a long short-term memory (LSTM) based network receives a layer-by-layer encoding. In contrast to our approach, they  encode an epoch number and predict the accuracy at the given epoch. Peephole delivers good performance on MNIST and CIFAR-10, however it has not been designed to transfer knowledge from familiar datasets to unseen ones. Given a new dataset, hundreds of networks need to be trained before Peephole makes a prediction. In contrast, our framework is designed to operate on unseen datasets, without the need of expensive training.

Accuracy predictors such as learning curves extrapolation (LCE)~\cite{lce}, BNN~\cite{bnn}, $\nu$-SVR~\cite{svr} forecast network performance based on partial learning curves. These algorithms are designed in the context of hyperparameter optimization or meta-learning. Both cases require extensive use of training and thus result in high computational costs. Moreover, they are all dataset and network specific, i.e., the prediction cannot be transferred to another network or dataset, without re-training. In particular, LCE employs a weighted probabilistic model to predict network performance. BNN uses Bayesian Neural Networks to fit completely new learning curves and extrapolate partially observed ones. This approach yields superior performance compared to~LCE, particularly at stages where the initial observed learning curve is not sufficient for the parametric algorithm to converge. Nevertheless, both methods rely on expensive Markov~Chain Monte~Carlo sampling procedures.
$\nu$-SVR~\cite{svr} complements the information on the learning curve with network architecture details and a list of predefined hyperparameters. These are used to train a sequence of regression models, that outperform LCE and~BNN. Although these methods exhibit good performance, they require a considerable part of the initial learning curve to provide reliable performance.  

Large-scale exploration algorithms~\cite{real2017large,DBLP:journals/corr/XieY17,DBLP:journals/corr/MiikkulainenLMR17,DBLP:journals/corr/abs-1708-05552,DBLP:journals/corr/ZophL16,zoph2017learning,cai2018efficient,baker,pham2018efficient} employ genetic mutations or reinforcement learning to explore a large space of architecture configurations.
Regardless of the approach, all these methods train a large number of networks, some of them employing hundreds of GPUs for more than ten days~\cite{real2017large}. ENAS~\cite{pham2018efficient} uses a controller to discover CNN architectures, by searching for an optimal subgraph within a large computational graph. With this approach it discovers a $97.11\%$ accurate network for CIFAR-10, on a single GPU in 10~hours. While ENAS reduces drastically the time-to-solution compared to previous results, the model is applied to only one dataset and not generalized to the case of multiple datasets. Indeed, sharing parameters among child models for different datasets is not straightforward.

\section{Methodology} \label{sec:methodology}

\begin{figure}[!t]
\includegraphics[width=\textwidth]{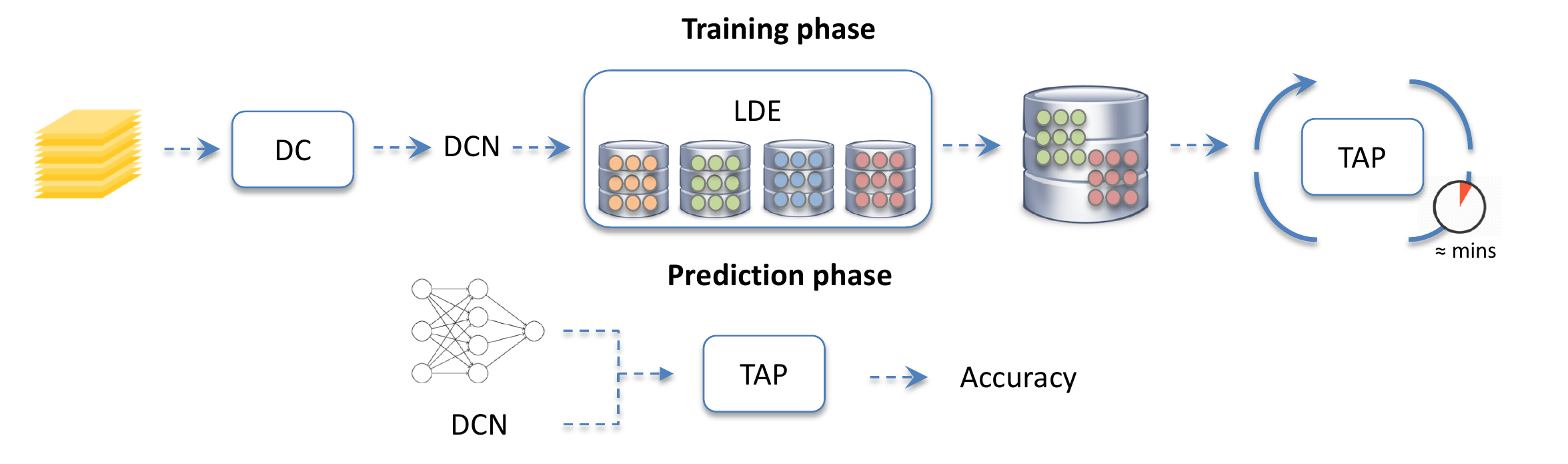}
\caption{Schematic \TAPAS workflow. First row: the Dataset Characterization (\DCC) takes a new, unseen dataset and characterizes its difficulty by computing the Dataset Characterization Number (\DCN). This number is then used to select a subset of experiments executed on similarly difficult datasets from the Lifelong Database of Experiments (\LCDB). Subsequently, the filtered experiments are used to train the Train-less Accuracy Predictor (\AP), an operation that takes up to a few minutes. Second row: the trained \AP takes the network architecture structure and the dataset \DCN and predict the peak accuracy reachable after training. This phase scales very efficiently in a few seconds over a large number of networks.}
\vspace*{-1em}
\label{fig:tapas_workflow}
\end{figure}

In this section, we provide a detailed overview of the main building blocks of the \TAPAS framework. \TAPAS aims to reliably estimate peak accuracy at low cost  for a variety of CNN architectures. This is achieved by leveraging a compact characterization of the user-provided input dataset, as well as a dynamically growing database of trained neural networks and associated performance.
The \TAPAS framework, depicted in Figure~\ref{fig:tapas_workflow}, is built on three main components:

\begin{enumerate}
\item \textbf{Dataset Characterization (\DCC):}
Receives an unseen dataset and computes a scalar score, namely the Dataset Characterization Number (\DCN)~\cite{dataset_char}, which is used to rank datasets;

\item \textbf{Lifelong Database of Experiments (\LCDB):} 
Ingests training experiments of NNs on a variety of image classification datasets executed inside the \TAPAS framework;

\item \textbf{Train-less Accuracy Predictor (\AP):} Given an NN architecture and a \DCN, it predicts the potentially reachable peak accuracy without training the network.
\end{enumerate}

\subsection{Dataset characterization (\DCC)} \label{sec:dataset_characterization}

The same CNN can yield different results if trained on an easy dataset (e.g., MNIST~\cite{data_mnist}) or on a more challenging one (e.g., CIFAR-100~\cite{data_cifar10_100}), although the two datasets might share features such as number of classes, number of images, and resolution. Therefore, in order to reliably estimate a CNN performance on a dataset we argue that we must first analyze the dataset difficulty. We compute the \DCN by training a \emph{probe net} to obtain a dataset difficulty estimation~\cite{dataset_char}. We use the \DCN for filtering datasets from the \LCDB and directly as input score in the \AP training and prediction phases as described in Section~\ref{sec:ap}.

\subsubsection{\DCN computation}

\emph{Prob nets} are modest-sized neural networks designed to characterize the difficulty of an image classification dataset~\cite{dataset_char}. We compute the \DCN as peak accuracy, ranged in $[0, 1]$, obtained by training the \textit{Deep normalized ProbeNet} on a specific dataset for ten epochs. The \DCN calculation cost is low due the following reasons: (i)~\textit{Deep norm ProbeNet} is a modest-size network, (ii)~the characterization step is performed only once at the entry of the dataset in the framework (the \LCDB stores the \DCN afterwards), (iii)~the \DCN does not require an extremely accurate training, thus reducing the cost to a few epochs, and (iv)~large datasets can be subsampled both in terms of number of images and of pixels.

The \DCN is a rough estimation of the dataset difficulty, and is thus tolerant to approximations. In Section~\ref{sec:experiments} we provide evidence of the effect of the \DCN on the \AP.

\subsection{Lifelong database of experiments (\LCDB)}\label{sec:LDE}

\LCDB is a continuously growing DB, which ingests every new experiment effectuated inside the framework. An experiment includes the CNN architecture description, the training hyper-parameters, the employed dataset (with its \DCN), as well as the achieved accuracy. 

\subsubsection{\LCDB initialization}\label{sec:lde_initialization}

At the very beginning, the \LCDB is empty. Thus we perform a massive initialization procedure to populate it with experiments. 
For each available dataset in Figure~\ref{fig:datasets} we sample 800~networks from a slight variation of the space of MetaQNN~\cite{baker}. For convolution layers we use strides with values in $\{1, 2\}$, receptive fields with values in $\{3, 4,.. 256\}$, padding in $\{same, valid\}$ and whether is batch normalized or not. We also add two more layer types to the search space: residual blocks and skip connections. The hyperparameters of the residual blocks are the receptive field, stride and the repeat factor. The receptive field and the stride have the same bounds as in the convolution layer, while the repeat factor varies between 1 and 6 inclusively. The skip connection has only one hyperparameter, namely the previous layer to be connected to.

To speed up the process, we train the networks one layer at a time using the incremental method described in~\cite{incremental}. In this way we obtain the accuracies of all intermediary sub-networks at the same cost of the entire one. To facilitate the \AP, we train all networks with the same hyper-parameters, i.e., same optimizer, learning rate, batch size, and weights initiallizer. Although the fixed hyper-parameter setting seems a strong limitation and might limit peak accuracy by a few percent, it is enough to trim poorly performing networks and, in the case of an architecture search, to fairly rank competitive networks, the performance of which can later be  optimized further, as discussed in Section~\ref{sec:experiments}. As data augmentation we use standard horizontal flips, when possible, and left/right shifts with four pixels. For all datasets we perform feature-wise standardization.

\subsubsection{\LCDB selection}\label{sec:lde_prefiltering}

Let us consider an \LCDB populated with experiments from $N_d$~different datasets $D_j$, with $j=1,\dots,N_d$. Given a new input dataset~$\hat{D}$ and its corresponding characterization $\textrm{\DCN}(\hat{D})$, the \LCDB block returns all experiments performed with datasets that satisfy the following relation
\begin{equation}
\| \textrm{\DCN}(\hat{D}) - \textrm{\DCN}(D_j) \| \leq \tau \qquad j \in [1,N_d],
\label{eq:florian_DCN}
\end{equation}
where $\tau$ is a predefined threshold that, in our experiments, is set to 0.05.

\subsection{Train-less accuracy predictor (\AP)}\label{sec:ap}

\AP is designed to perform fast and reliable CNN accuracy predictions. Compared to Peephole~\cite{peephole}, \AP leverages knowledge accumulated through experiments of datasets of similar difficulty filtered from the \LCDB based on the \DCN. Additionally, \AP does not first analyze the entire NN structure and then makes a prediction, but instead performs an iterative prediction as depicted in Figure~\ref{fig:encoding}. In other words, it aims to predict the accuracy of a sub-network $l_{1:i+1}$, assuming the accuracy of the sub-network $l_{1:i}$ is known. The main building elements of the predictor are: (i) a compact encoding vector that represents the main network characteristics, (ii) a quickly-trainable network of LSTMs, and (iii) a layer-by-layer prediction mechanism. 

\subsubsection{Neural network architecture encoding} \label{sec:network_encoding}

\begin{figure}[!t]
\includegraphics[width=\textwidth]{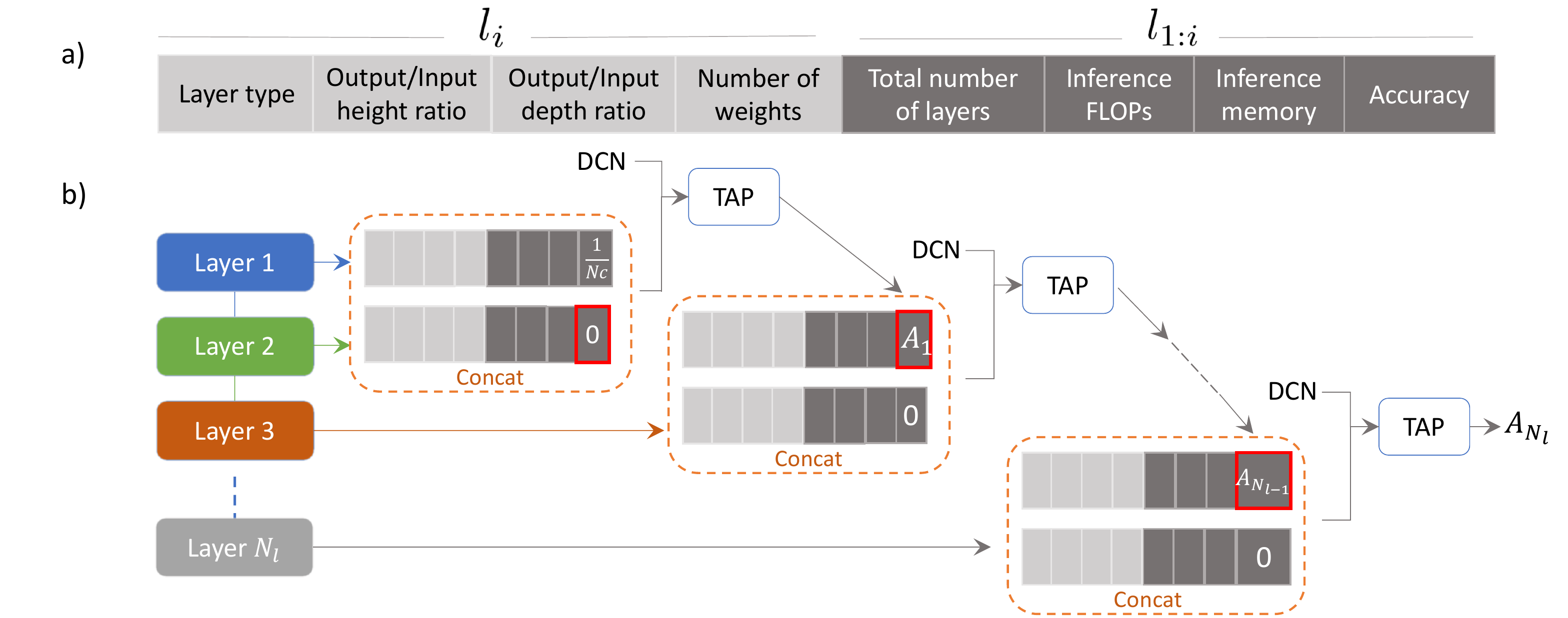}
\caption{Encoding vector structure and its usage in the iterative prediction. a) The encoding vector contains two blocks: $i$-layer information and from input to $i$-layer sub-network information. b) The encoding vector is used by the \AP following an iterative scheme. Starting from Layer~1 (input) we encode and concatenate two layers at a time and feed them to the \AP. In the concatenated vector, the \emph{Accuracy} field $\textrm{A}_i$ of $l_i$ is set to the predicted accuracy obtained from the previous \AP evaluation, whereas the one of $A_{i+1}$ corresponding to $l_{i+1}$ is always set to zero. For the input layer, we set $\textrm{A}_0$ to $1/N_c$, where $N_c$ is the number of classes, assuming a random distribution. The final predicted accuracy $A_{N_l}$ is the accuracy of the complete network.}
\label{fig:encoding}
\vspace*{-1.0em}
\end{figure}

Similar to Peephole, \AP employs a layer-by-layer encoding vector as described in Figure~\ref{fig:encoding}. Unlike Peephole, we encode more complex information of the network architecture for a better prediction.

Let us consider a network with $N_l$~layers, $l_i$ being the $i$-th layer counting from the input, with $i=1,\dots,N_l$. We define a CNN sub-network as $l_{a:b}$ with $1 \le a < b \le N_l$. Our encoding vector contains two types of information as depicted in Figure~\ref{fig:encoding}~a): (i)~$i$-th layer information and (ii)~$l_{1:i}$~sub-network information. For the current $i$-th layer we make the following selection of parameters: \emph{Layer type} is a one-hot encoding that identifies either convolution, pooling, batch normalization, dropout, residual block, skip connection, or fully connected. Note that for the shortcut connection of the residual block we use both the identity and the projection shortcuts~\cite{resnet}. The projection is employed only when the residual block decreases the number of filters as compared to the previous layer. Moreover, as compared to~\cite{peephole}, our networks do not follow a fixed skeleton in the convolutional pipeline, allowing for more generality. We only force a fixed block at the end, by using a global pooling and a fully connected layer to prevent networks from overfitting~\cite{nin}.

The \emph{ratio between the output height and input height} of each layer accounts for different strides or paddings, whereas the \emph{ratio between the output depth and input depth} accounts for modifications of the number of kernels. The \textit{number of weights} specifies the total of learnable parameters in $l_i$. This value helps the \AP differentiate between layers that increase the learning power of the network (e.g., convolution, fully connected layers) and layers that reduce the dimensionality or avoid overfitting (e.g., pooling, dropout). In the second part of the encoding vector, we include: \emph{Total number of layers}, counting from input to $l_i$, \emph{Inference FLOPs} and \emph{Inference memory} that are an accurate estimate of the computational cost and memory requirements of the sub-network, and finally \emph{Accuracy}, which is set either to 1/$N_c$, for the first layer, where $N_c$ is the number of classes to predict, zero for prediction purposes, or a specific value $\mathrm{A}_i \in [0, 1]$ that is obtained from the previous layer prediction. Before training, we perform a feature-wise standardization of the data, meaning that for each feature of the encoding vector, we subtract the mean and divide by the standard deviation.

\subsubsection{\AP architecture}

\AP is a neural network consisting of two stacked LSTMs of 50 and 100 hidden units, respectively, followed by a single-output fully connected layer with sigmoid activation. The \AP network has two inputs. The first input is a concatenation of two encoding vectors corresponding to layer $l_i$ and $l_{i+1}$, respectively. This input is fed into the first LSTM. The second input is the \DCN and is concatenated with the output of the second LSTM and then fed into the fully connected layer. 

\subsubsection{\AP training}\label{sec:training}

\AP requires a significant amount of training data to make reliable predictions. The \LCDB provides this data as described in Section~\ref{sec:LDE}. As mentioned above, all our generated networks are trained in an incremental fashion, as presented in \cite{incremental}, meaning that for each network of length $N_l$ we train all intermediary sub-networks $l_{1:k}$ with $1 < k \leq N_l$ and save their performance $A_{k}$. We encode each set of two consecutive layers $l_i$ and $l_{i+1}$ following the schema detailed in \ref{sec:network_encoding}, setting the accuracy field in the encoding vector of $l_i$ to $A_i$, which was obtained through training, and aiming to predict $A_{i+1}$.

\AP is trained with RMSprop~\cite{rmsprop}, using a learning rate of $10^{-3}$, a HeNormal weight initialization~\cite{heinit}, and a batch size of 512. As the architecture of the \AP is very small, the training process is of the order of a few minutes on a single GPU device. Moreover, the trained \AP can be stored and reapplied to other datasets with similar DCN numbers without the need for retraining.

\subsubsection{\AP prediction}\label{sec:tap_prediction}

\AP employs a layer-by-layer prediction mechanism. The accuracy~$\textrm{A}_{i}$ of the sub-network~$l_{1:i}$ predicted by the previous \AP evaluation is subsequently fed as input into the next \AP evaluation, which returns the predicted accuracy~$\textrm{A}_{i+1}$ of the sub-network~$l_{1:i+1}$. This mechanism is described more in detail in Figure~\ref{fig:encoding} b).

\section{Experiments} \label{sec:experiments}
In this section, we demonstrate \TAPAS performance over a wide range of experiments. Results are compared with reference works from the literature.
All runs involve single-precision arithmetic and are performed on 
IBM\footnote{\fontsize{8}{6}\selectfont IBM, the IBM logo, ibm.com, OpenPOWER are trademarks or registered trademarks of International Business Machines Corporation in the United States, other countries, or both. Other product and service names might be trademarks of IBM or other companies.} POWER8 compute nodes, equipped with four NVIDIA~P100 GPUs.

\subsection{Dataset selection for \LCDB initialization}

\begin{figure}[!t]
\centering
\includegraphics[width=0.7\textwidth]{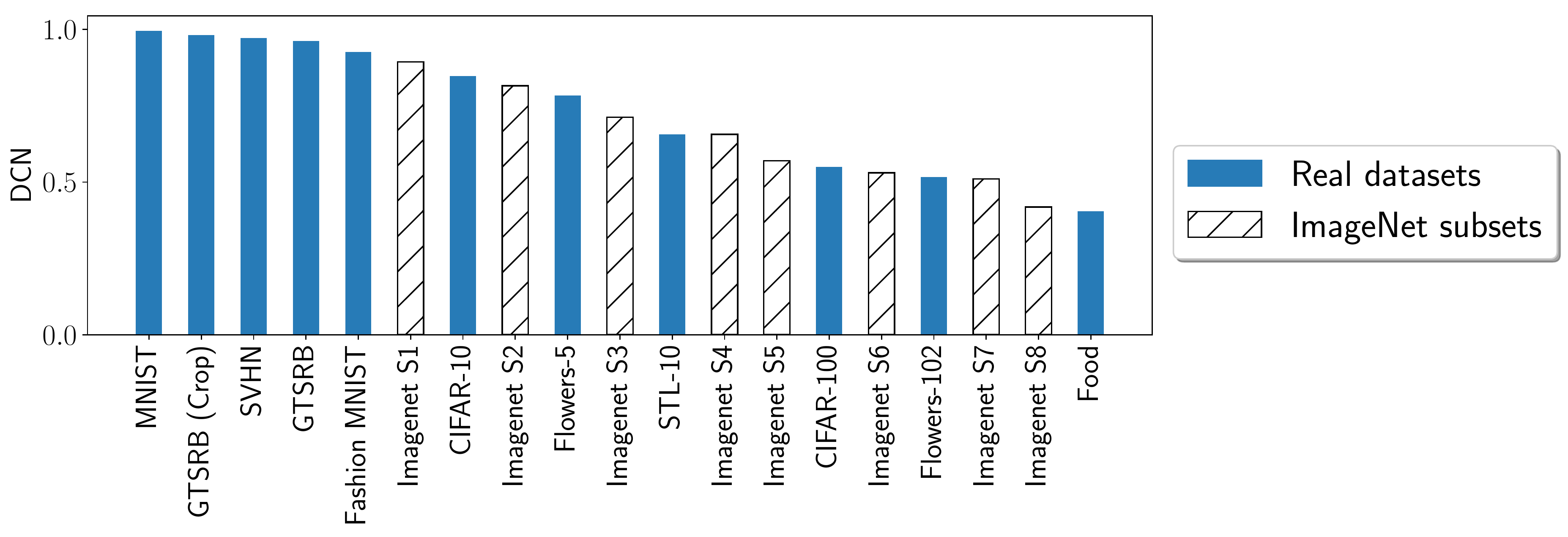}
\caption{List of image classification datasets used for characterization. The datasets are sorted by the \DCN value from the easiest (left) to the hardest (right).
}
\label{fig:datasets}
\vspace*{-1em}
\end{figure}

All the experiments are based on a \LCDB populated with nineteen datasets, ranked by difficulty in Figure~\ref{fig:datasets}. Eleven of them are publicly available. The other eight are generated by sub-sampling the ImageNet dataset~\cite{imagenet_cvpr09} varying the number of classes and the number of images per class. The result is a finer distribution of datasets per \DCN value, that improves the predictions by biasing \AP closer to the relevant data, particularly for the leave-one-out cross-validation experiment presented later. Additional details are provided in the Appendix.
We resize all dataset images to $32 \times 32$ pixels. On the one hand, this reduces the cost of \LCDB initialization, on the other hand, it allows us to potentially test networks and datasets in an \emph{All2All} fashion. We remark that this choice does not lead to a loss of generality, as images of different sizes can be employed in the same pipeline.
For every dataset, we generate 800~networks based on the procedure described in Section~\ref{sec:lde_initialization}. All networks are trained under the same settings: RMSprop optimizer with a learning rate of $10^{-3}$, weight decay $10^{-4}$, batch size 64, and HeNormal weight initialization.

\subsection{\TAPAS performance evaluation} \label{sec:experiments_predictor}

In this section, we define three different scenarios to compare \TAPAS with LCE~\cite{lce}, BNN~\cite{bnn}, $\nu$-SVR~\cite{svr} and Peephole~\cite{peephole}. We employ three evaluation metrics: (i)~the \emph{mean squared error (MSE)}, which measures the difference between the estimator and what is estimated, (ii)~\emph{Kendall's Tau (Tau)}, which measures the similarities of the ordering between the predictions and the real values, and (iii)~the \emph{coefficient of determination} ($R^2$), which measures the proportion of the variance in the dependent variable that is predictable from the independent variable. In the first metric, lower is better (zero is best); in the others, higher is better (one is best).

\subsubsection{Scenario A: Prediction based on experiments on a single dataset} 

\begin{figure}[!t]
\centering
\includegraphics[width=\textwidth]{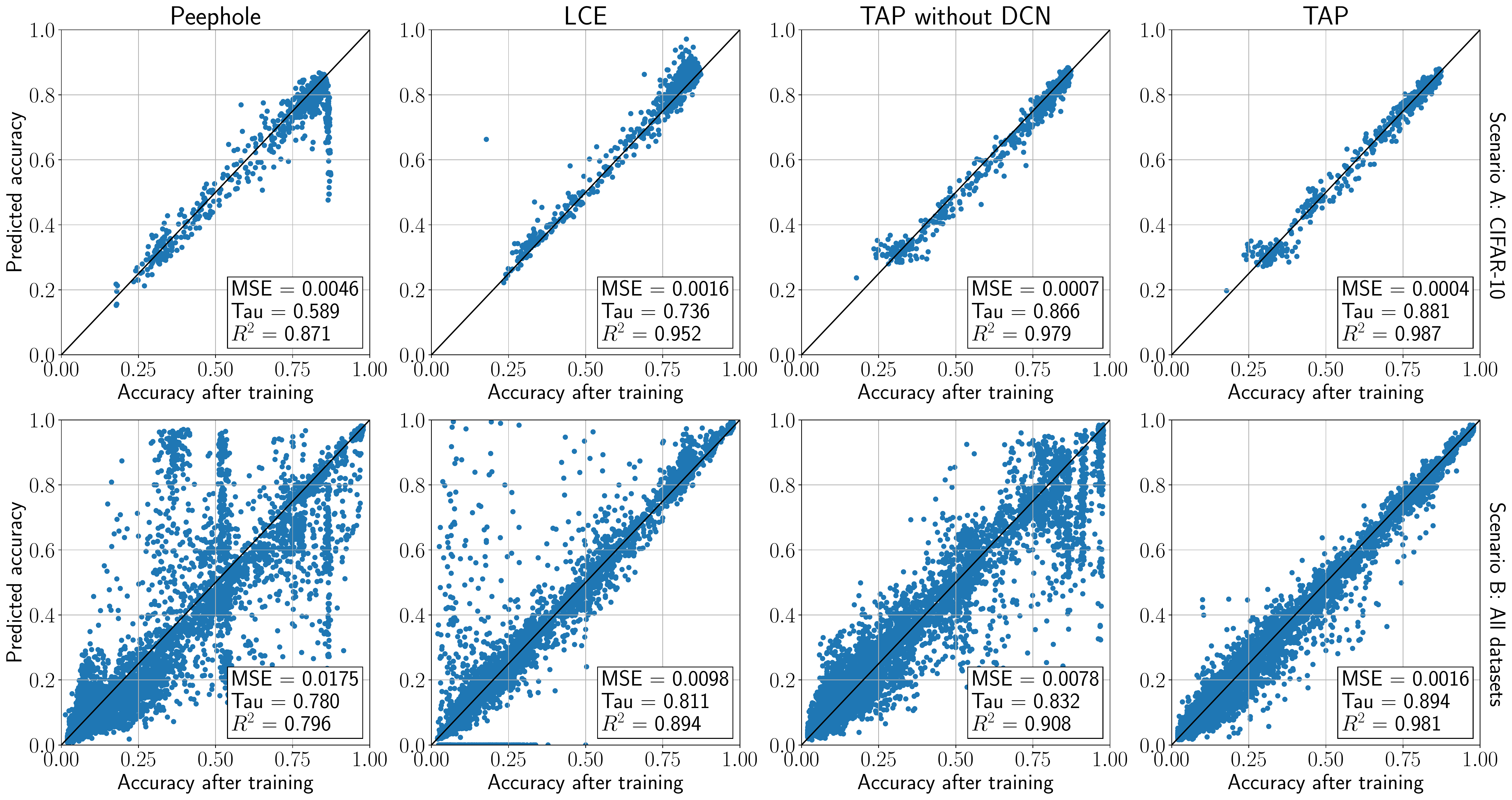}
\caption{Superior predictive performance of \AP compared with state-of-the-art methods, both when trained on only one dataset (Scenario A) or on multiple datasets (Scenario B).}
\label{fig:scenarios_a_and_b}
\vspace*{-1em}
\end{figure}

We train the \AP on a filtered list of experiments from the \LCDB based on the CIFAR-10 dataset. We recognize that this scenario is very favorable for prediction, however it is used in reference publications, and therefore allows for a fair comparison.

We perform ten-fold cross validation and present the results for Peephole, LCE, and \TAPAS in the first row of Figure~\ref{fig:scenarios_a_and_b}. For BNN and $\nu$-SVR we rely on published numbers. When presented with 20\% of the initial learning curve, BNN states an MSE of 0.007, while $\nu$-SVR states an $R^2$ of 0.9. \AP outperforms all methods, in terms of all the considered metrics. Moreover, if we modify \AP to not use the \DCN, we still get better predictions than with all the other methods. The \AP prediction performance is not strongly affected because the training and prediction involve only one dataset. 

We argue that the lower results of the Peephole method, as compared to the original paper, are due to the more complicated structure of the network we used in our benchmark. Specifically, the Peephole-encoding tuple (layer type, kernel height, kernel width, channels ratio) is not sufficient to predict complicated structures like ResNets.

\subsubsection{Scenario B: Prediction based on experiments on all datasets}

This scenario is similar to Scenario~A, but we do not filter experiments by dataset. The second row of Figure~\ref{fig:scenarios_a_and_b} shows results when \AP is trained on all datasets, regardless of their \DCN. Also in this scenario, \AP outperforms all methods in all of the considered metrics. We recognize that Peephole is designed to be dataset-specific. However, compared to \AP without \DCN the comparison is fair, as neither of these algorithms contain information about the dataset difficulty.

\subsubsection{Scenario C: Prediction based on experiments on unseen datasets}

\begin{figure}
\centering
\includegraphics[width=0.9\textwidth]{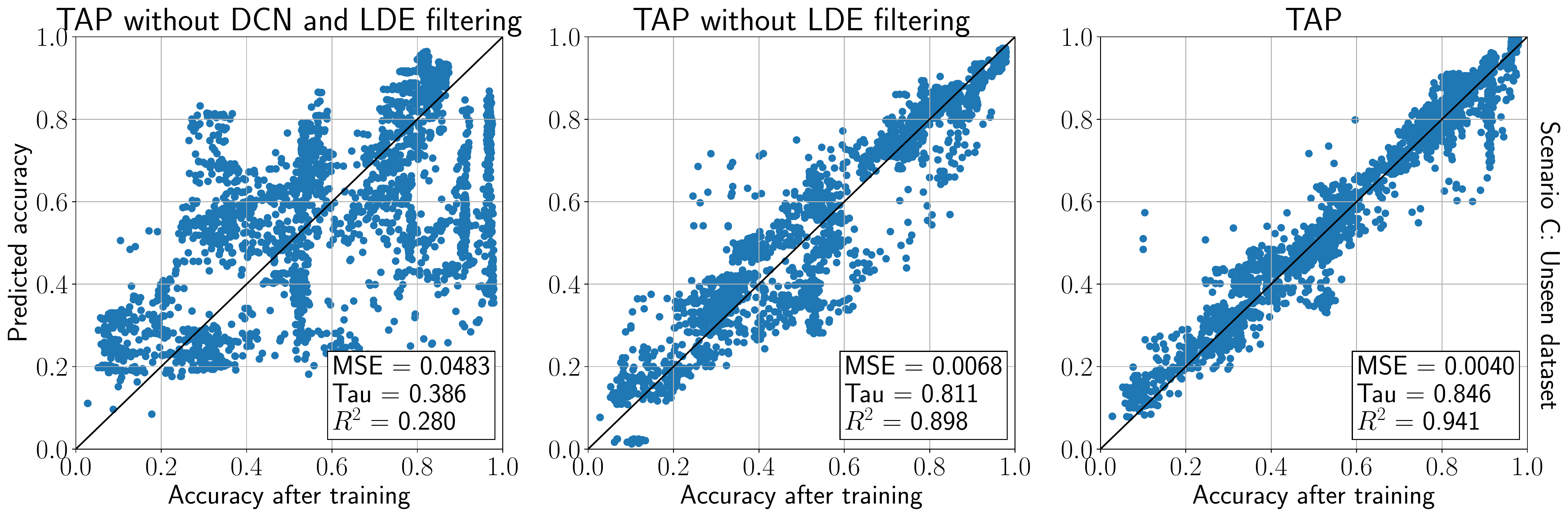}
\caption{Predicted vs real performance (i.e., after training) for Scenario~C. Left plot: \AP trained without \DCN or \LCDB pre-filtering. Middle plot: \AP trained with \DCN, but \LCDB is not pre-filtered. Right plot: \AP trained only on \LCDB experiments with similar dataset difficulty, according to \eqref{eq:florian_DCN}.} 
\label{fig:scenario_c}
\vspace*{-0.7em}
\end{figure}

This scenario aims (i)~to demonstrate \TAPAS performance when targeting completely unseen datasets and (ii)~to highlight importance of dataset-difficulty characterization and \LCDB pre-filtering. To do that, we consider the list of datasets in Figure~\ref{fig:datasets} and perform eleven leave-one-out cross-validation benchmarks, considering only the real datasets. 
The result of this experiment is presented in Figure~\ref{fig:scenario_c}. From left to right, we observe the cumulative impact of the \DCN awareness in the \AP training, as well as of the pre-filtering of the experiments in the \LCDB according to~\eqref{eq:florian_DCN}. Moreover, by comparing the rightmost plot and metrics with previous results in Figure~\ref{fig:scenarios_a_and_b}, we observe that \TAPAS performance does not diminish significantly when applied to an unknown dataset.

\subsection{Simulated large-scale evolution of image classifiers} \label{sec:search}

The \AP can be plugged into any large-scale evolution algorithm to perform train-less architecture search. In this work, we use the genetic algorithm introduced in~\cite{real2017large}. As described in the original paper, the evolution algorithm begins with a small population, consisting of one~thousand single-layered networks. After training, two candidates are randomly chosen from the population: the less accurate one is removed, whereas the other one undergoes a mutation. The mutated network is evaluated in roughly 30 epochs and then put back in the population. The operation repeats until convergence is achieved. 

The above algorithm is very expensive: 250~parallel workers are used for training the population and the entire process takes 256~hours~\cite[Figure~1]{real2017large}. The \AP can simulate the large-scale evolution search in only 400~seconds on a single GPU~device performing 20k~mutations. We employ the same mutations as in~\cite{real2017large}, apart from those that do not make sense in a simulation, such as altering the learning rate and resetting the weights. No network is trained during the entire process.

\begin{figure}
\centering
\begin{subfigure}{.5\textwidth}
  \centering
  \includegraphics[width=\linewidth]{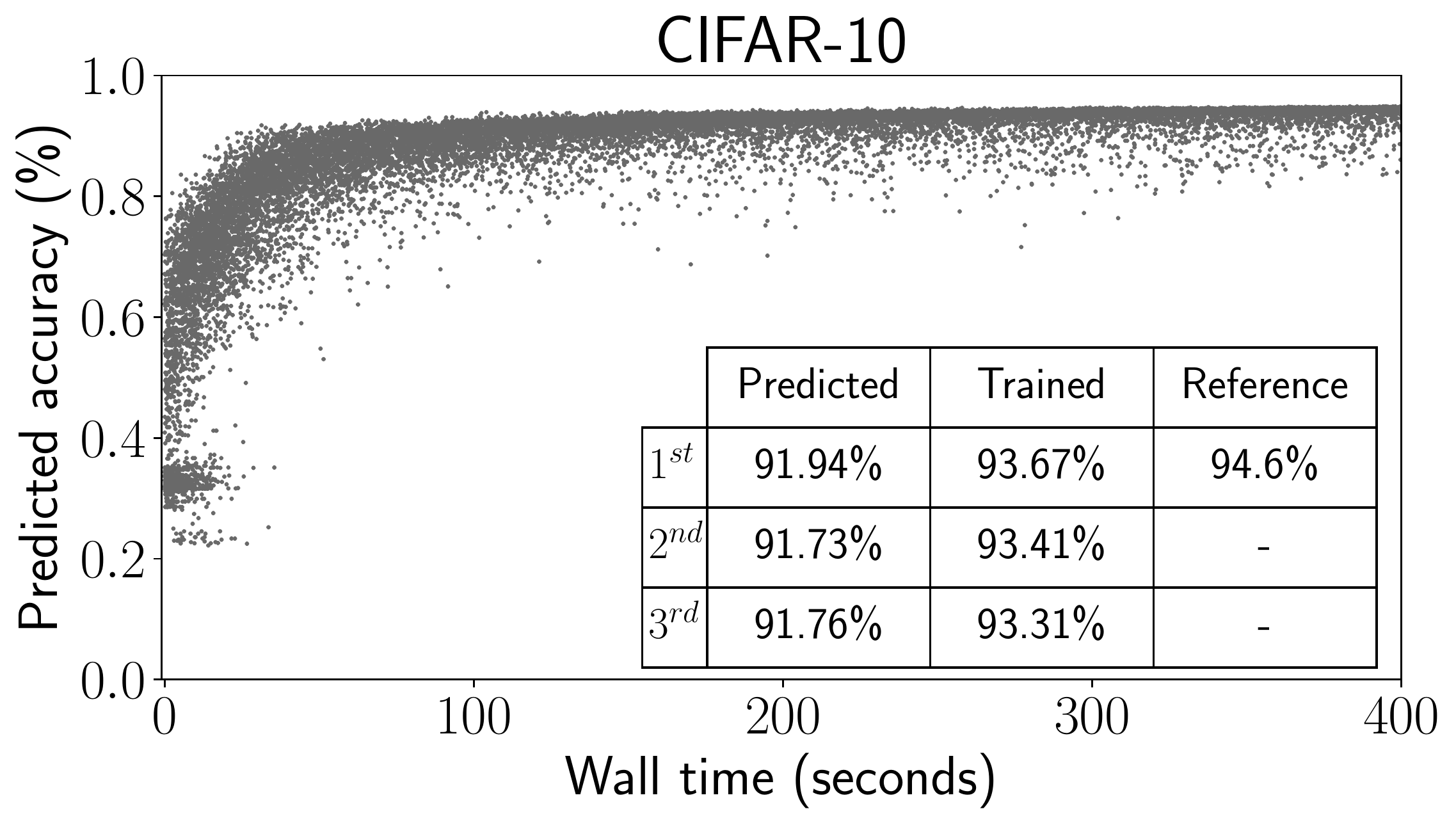}
  \label{fig:sub1}
\end{subfigure}%
\begin{subfigure}{.5\textwidth}
  \centering
  \includegraphics[width=\linewidth]{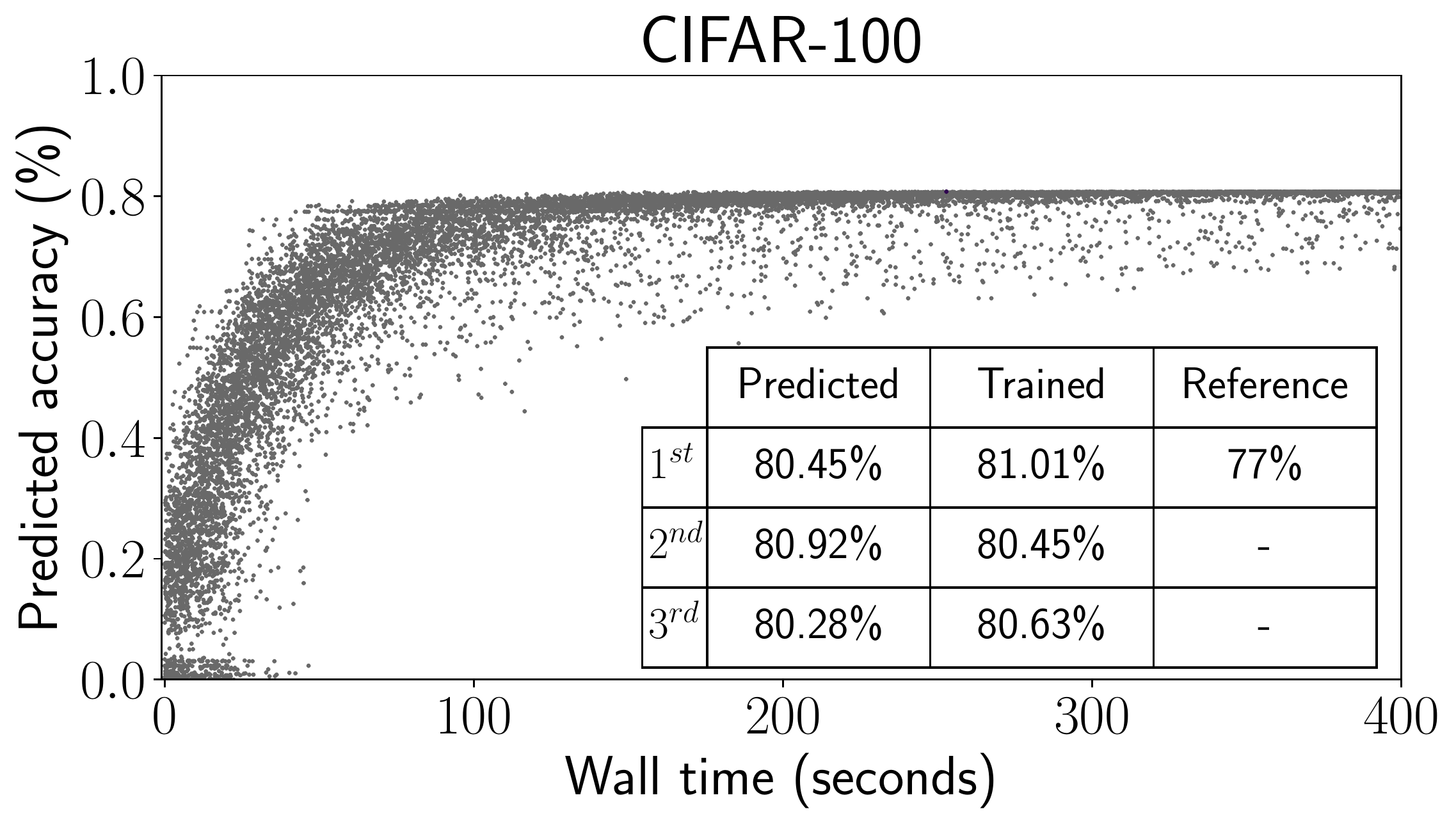}
  \label{fig:sub2}
\end{subfigure}
\vspace*{-0.7em}
\caption{Simulation of large-scale evolution, with 20k~mutations. The table compares top three networks (predicted and trained accuracy) with reference work~\cite{real2017large}. The simulations require only $2\times10^{11}$ FLOPs per dataset, while training the top-three networks for 100 epochs is an additional $3\times10^{15}$ FLOPs, causing a 6 hour runtime on a single GPU. The reference work employs $9\times10^{19}$~(CIFAR-10) and $2\times10^{20}$ FLOPs (CIFAR-100) causing a runtime of 256 hours on 250 GPUs.}
\label{fig:large_scale_evolution}
\vspace*{-1em}
\end{figure}

Figure~\ref{fig:large_scale_evolution} presents results of the simulated evolution for both CIFAR-10 and CIFAR-100 datasets. To verify that the \AP discovers good networks, we select the top three networks (according to accuracy prediction) and train them a-posteriori. For CIFAR-10 our best network reaches 93.67\%, whereas for CIFAR-100 we achieve 81.01\%, an improvement of 4\% w.r.t.\ the reference work~\cite{real2017large}. Moreover, we observe that all the top three networks perform well, and prediction values are reasonably close to those after training.

\section{Conclusion}\label{sec:conclusion}
In this paper we propose \TAPAS, a novel prediction framework that given a CNN architecture, accurately forecasts its performance at convergence (i.e., peak validation accuracy) for any given input dataset. \TAPAS know-how originates from a lifelong database of experiments, based on a wide variety of datasets. Reliance on dataset-\emph{difficulty} characterization, is our key differentiation to outperform state-of-the-art methods by a large margin. Indeed, we demonstrated that \TAPAS outperforms preexisting methods, both in the favourable case when the methods are tuned for a specific dataset, as well as when they are applied on a wide range of datasets, without any bias.
\TAPAS does not require new training experiments, even in the  case scenario when it is applied to a completely new dataset. 
This facilitates large-scale network architecture searches, that do not require executions of training jobs. Indeed, \TAPAS enabled us to identify very accurate CNN architectures, in a few minutes, using only  a single GPU. This is a performance that is several orders of magnitude faster than any training-based approach.

\small
\bibliography{citations.bib}

\end{document}